\documentclass[12pt]{article}
\usepackage{apacite}
\usepackage{times}

\usepackage[english]{babel}
\usepackage{csquotes}
\usepackage{lipsum}
\usepackage{setspace}

\usepackage{color}
\usepackage[tmargin=1in,bmargin=1in,lmargin=1.25in,rmargin=1.25in]{geometry}
\usepackage{subfig}
\usepackage{multirow}
\usepackage{booktabs}
\usepackage{graphicx}
\usepackage{engord}

\usepackage{tabularx}

\usepackage[most]{tcolorbox}

\title{Harnessing Artificial Intelligence to Combat Online Hate: Exploring the Challenges and Opportunities of Large Language Models in Hate Speech Detection}
\author{Tharindu Kumarage$^*$~ Amrita Bhattacharjee\thanks{Equal Contribution.} ~ Joshua Garland \\ Arizona State University \\
        \texttt{\small \{kskumara,abhatt43,jtgarlan\}@asu.edu} }

\date{}

\begin{document}

\maketitle
\begin{abstract}
Large language models (LLMs) excel in many diverse applications beyond language generation, e.g., translation, summarization, and sentiment analysis. One intriguing application is in text classification. This becomes pertinent in the realm of identifying hateful or toxic speech--- a domain fraught with challenges and ethical dilemmas. In our study, we have two objectives: firstly, to offer a literature review revolving around LLMs as classifiers, emphasizing their role in detecting and classifying hateful or toxic content. Subsequently, we explore the efficacy of several LLMs in classifying hate speech: identifying which LLMs excel in this task as well as their underlying attributes and training. Providing insight into the factors that contribute to an LLM’s proficiency (or lack thereof) in discerning hateful content. By combining a comprehensive literature review with an empirical analysis, our paper strives to shed light on the capabilities and constraints of LLMs in the crucial domain of hate speech detection. 
\end{abstract}

\maketitle
\newpage
\section{Introduction} 

Online social media platforms have become important channels of communication and sharing information, opinions, and connecting with other individuals and businesses. However, these platforms are also often used for hateful or toxic content, bullying and intimidation, etc.~\cite{poletto2021resources}. Given the scale of such platforms, hate speech and toxic content detection is a challenge and performing such detection manually is infeasible. This necessitates the use of automated detection systems~\citeA{del2017hate,schmidt2017survey}, which also is a challenge in practice due to the dynamic nature of hate speech~\citeA{sheth2023peace}. Hate speech can evolve with time, is highly subjective, and may be dependent on the context in which it is expressed~\citeA{macavaney2019hate, sheth2024causality}.

With the advent of advanced large language models (LLMs), there is growing interest in leveraging these models for content moderation. Specifically, using them to detect harmful and toxic content online by simply prompting the models. Several recent studies have examined the efficacy of GPT-3~\citeA{brown2020language} and GPT-3.5~\citeA{huang2023chatgpt}\footnote{ChatGPT and GPT-3.5 are used interchangeably here.} in detecting hate speech, encompassing both explicit and implicit forms. OpenAI has recently presented in-house experiments demonstrating GPT-4's~\citeA{openai2023gpt} potential as a content moderator\footnote{https://openai.com/blog/using-gpt-4-for-content-moderation}. Similarly, the state-of-the-art open-source model, Llama 2~\citeA{touvron2023llama}, has shown promise in hate speech detection. In this study, our objective is to thoroughly assess these claims and delve into the nuances behind the LLMs' ability to discern hate speech. To achieve this, first explore the space of LLMs as a detector or text classifier, with a focus on the task of hate speech detection. Then, we evaluate several candidate LLMs, spanning both open-source and proprietary models, and address the following research questions:

Q1: How robust are these LLMs in detecting hate speech?
We will examine and compare multiple LLMs on various types of hate speech: both general and targeted towards specific minorities. We aim to determine if these LLMs primarily rely on specific keywords, such as profanities, for detection, or if they genuinely discern and characterize the hateful intent of the speech. 

Q2: How do various prompting techniques influence the hate speech detection efficacy of LLMs?
We will compare different prompting strategies, with varying degrees of complexity, to discern differences in how they affect the hate speech detection capabilities.
Based on our findings, we will endeavor to provide insights into the specific elements and nuances of LLMs and best practices surrounding the use of LLMs for this particular task.

\section{LLMs as Text Classifiers or Annotators}

Given the availability of several large language models, both open-source and proprietary (albeit via APIs), these technologies are increasingly being used in NLP applications such as text classification. Owing to the success of the more recent larger LLMs (such as ChatGPT, GPT-4~\citeA{openai2023gpt}, Llama 2~\citeA{touvron2023llama}, etc.), researchers are actively exploring novel use-cases of such models in order to tackle issues such as generalization, data scarcity, etc. In this section we provide a brief overview on how language models (both pre-trained language models, and the more recent large language models) have been used in the task of text classification, first going over the general text classification task, before delving into hate speech specific classifiers.

\label{sec:LLM}

\subsection{General Text Classifier or Annotator}

In this section, we describe some works that have used language models for the general problem of text classification. We further divide this section into two categories: (i) the pre-LLM era, and (ii) the LLM era.

\subsubsection*{Pre-LLM Era}

In the pre-LLM era, pre-trained language models (PLMs) such as BERT ~\citeA{devlin2018bert}, RoBERTa~\citeA{liu2019roberta}, BART~\citeA{lewis2019bart} etc. have been used extensively as language encoders. These PLMs are essentially transformer-based language models that are pre-trained on a large corpus of unlabeled text data (mostly webtext) and often fine-tuned on downstream task datasets to perform classification or detection. Given the extensive pre-training that these language models go through, PLMs are often used as general language encoders in a classification task, with additional classification layers or classification heads added to facilitate task-specific fine-tuning ~\citeA{howard2018universal,arslan2021comparison}.

For example, authors in ~\citeA{kant2018practical} first pre-train and then fine-tune an encoder-decoder type language model on task specific data for the task of multi-dimensional sentiment classification and compare their method with a pre-trained ELMo~\citeA{peters1802deep}, which is then further fine-tuned on their tasks-specific dataset. BERT~\citeA{devlin2018bert}, which is a bidirectional transformer-based language model, has shown impressive performance on many natural language understanding tasks. Authors in ~\citeA{sun2019fine} investigate the training regimes and different fine-tuning settings to understand how to get the most out of fine-tuning BERT for the task of text-classification. Through their experiments they advise that text classification using BERT can be improved via the following best practices: further pre-training on task-specific in-domain data, multi-task fine-tuning rather than single task fine-tuning etc. 

Given the smaller sizes of pre-trained language models as compared to more recent models like ChatGPT or Llama, these models have been used in several other text classification tasks, often with task-specific fine-tuning or in conjunction with other specialized architecture or training regimes~\citeA{min2023recent}. Examples of some tasks where such pre-trained language models have been used are toxic comment classification~\citeA{zhao2021comparative}, counter-speech detection~\citeA{garlandCounterClassifier,garland2020impact} text mining~\citeA{zhang2021smedbert}, sentiment classification~\citeA{meng2020text,rathnayake2022adapter}, etc.

\subsubsection*{LLM Era}

Given the impressive performance of newer LLMs such as ChatGPT and GPT-4~\citeA{openai2023gpt} on a variety of natural language tasks, that too in a zero-shot manner, researchers are evaluating the possibility of using such LLMs as annotators. This could potentially assuage data scarcity issues in tasks and thereby facilitate or improve training of better models. One recent work~\citeA{gilardi2023chatgpt} performed a systematic evaluation of the annotation capabilities of ChatGPT especially in comparison to annotations obtained from crowd workers on Amazon Mechanical Turk\footnote{https://www.mturk.com/}. They evaluate the accuracy of ChatGPT and MTurk workers with annotations from trained annotators and show that ChatGPT outperforms the MTurk crowd workers, on a variety of content moderation tasks involving different four datasets of Tweets and news articles. 

Another recent study~\citeA{zhu2023can} evaluated the capability of ChatGPT to reproduce human-generated labels on a set of five benchmark text datasets, on tasks such as stance detection, bot detection, sentiment analysis and hate speech detection. Results show that ChatGPT can replicate the human generated labels to a certain extent, achieving an accuracy of 0.609 across the five datasets, but is still far from being a perfect annotator. The authors also find varying performance of ChatGPT across different labels within one specific task. A similar observation has been made by authors in ~\citeA{bhattacharjee2023fighting} where ChatGPT was used to distinguish AI-generated text from human-written text, and an asymmetric performance across the two labels was identified. However, experiments demonstrate that GPT-4 has superior performance on the task. A similar work uses ChatGPT in automatic genre classification, where the task is to classify a given text into one of several genre categories such as News, Legal, Promotion, etc. The authors evaluate ChatGPT and compare its performance with a fine-tuned XLM-RoBERTa, and they test on both English and Slovenian language data. Interestingly, for the English split, ChatGPT performs better than the fine-tuned XLM-RoBERTa model, even without any labeled data, although the performance drops a bit for the Slovenian one. 

Compared to all these works that demonstrate the potential for using LLMs and, in particular ChatGPT as an annotator, one interesting piece of work~\citeA{reiss2023testing} investigates the reliability of ChatGPT-derived annotations, and demonstrates that the annotations rely heavily on the temperature parameters and possibly other factors such as length of the text prompt and complexity of instructions. 

\subsection{Hate Speech Classifiers}

In this section, we go over recent works that have used language models in a hate speech classification task, and we divide this section into two categories: (i) the pre-LLM era, and (ii) the LLM era.

\subsubsection*{Pre-LLM Era}

Similar to the general classification, early applications of language models in hate speech detection employed pre-trained language models as rich embeddings or representations for the text. Since hate speech detection is often heavily dependent on language-specific words and phrases such as profanities, there have been many efforts in building hate speech classifiers for specific languages. Among methods that use pre-trained language models in the detection framework, some examples are ~\citeA{plaza2021comparing} for Spanish hate speech detection where they use both multilingual pre-trained LMs like mBERT and XLM~\citeA{lample2019cross} as well as a Spanish version of BERT called BETO\footnote{https://github.com/dccuchile/beto}. Authors in ~\citeA{pham2020universal} build a detector for Vietnamese hate speech by using a RoBERTa model, or in particular, a version trained for the Vietnamese language called PhoBERT~\citeA{dat2020phobert}. Similar efforts involving detection using multilingual and monolingual versions of BERT or RoBERTa have also been done for Italian hate speech detection~\citeA{lavergne2020thenorth}, where alongside multilingual models, Italian versions such as AlBERTo, PoliBERT and UmBERTo have been used. Similar efforts for training language-specific hate speech detectors by fine-tuning different variants of the BERT family of models have been used in languages such as Marathi~\citeA{velankar2022mono}, Polish~\citeA{czapla2019universal}. 

Authors in ~\citeA{stappen2020cross} use frozen pre-trained language models as feature extractors in a framework for cross-lingual hate speech detection. Alongside comparing various framework designs for the task, authors also evaluate their proposed method in zero-shot and few-shot setting with substantial success. Another interesting work in multi-lingual hate speech detection uses a multi-channel BERT~\citeA{sohn2019mc}, i.e., multiple language-specific pre-trained BERT models in parallel to facilitate transfer learning, The authors also experiment with adding additional classification signals by providing translated versions of the input to the classifier. Given that the lack of labeled data in low-resource languages is a major bottleneck in the development of hate speech detectors for these particular languages, ~\citeA{zia2022improving} proposed a framework that leverages labeled data from a high-resource language such as English and used a language model based teacher-student framework to perform transfer learning for hate speech detection on a target language, in the absence of target labels. To do this, they first fine-tune a multilingual language model on labeled training data from the source language. Then they use this model to generate pseudo-labels for samples from the target language, by simply predicting in a zero-shot manner. Finally, they use these pseudo-labels to fine-tune a monolingual pre-trained language model to perform hate speech detection on the target language without requiring any labels from the target. 

\subsubsection*{LLM Era}

Most of the works discussed above use pre-trained language models of parameter sizes in the range of a few hundred million. However, there is a growing trend towards developing and training larger language models, often with parameter sizes of a few hundred billion. Performance of language models on NLP tasks have shown huge improvements with increase in the scale of these models. These larger models, now often referred to as Large Language Models (LLMs) are trained on huge internet-scale data corpora. Due to their extensive pre-training, LLMs often demonstrate good performance on a variety of tasks even on a zero-shot manner. The standard mode of using these LLMs is via the task of text generation, whereby the user provides a text input as a `prompt' to the LLM, and the LLM produces some text output conditioned on the input prompt. 

Broadly, there are two categories of LLMs: base LLMs - that simply perform the task of next token prediction, essentially performing a text completion task; and instruction-tuned LLMs - where LLMs are specifically trained to follow instructions in the prompt. Instruction-tuned LLMs are useful for a variety of tasks. Examples of such instruction-tuned LLMs are ChatGPT, GPT-4, the Llama family of models, etc. An example of a base LLM is GPT-3~\citeA{brown2020language} by OpenAI, with 175 billion parameters. 

Authors in ~\citeA{chiu2021detecting} evaluate the performance of GPT-3~\citeA{brown2020language} on hate speech detection in a variety of settings, including zero-shot, one-shot (where a single example is provided in the prompt), few-shot (where a small number of samples are provided in the prompt as examples). The authors also evaluate the few-shot performance along with instructions in the prompt wherein a small instruction is also provided in the prompt, specifying what the possible labels are, such as `sexist', `racist' or `neither'. Interestingly, the study finds that GPT-3 performs the best when prompted without instructions in a few-shot setting. In a similar direction, alongside experimenting with different prompt structures for this task, ~\citeA{han2022designing} shows how increasing the number of labeled samples in the prompt in the few shot setting improves the performance of GPT-3. 

Other recent prompt-based detection methods include ~\citeA{luo2023towards}, where the authors propose a new category of the hate speech detection task: enforceable hate speech detection, where text content is classified as hate speech if it violates at least one legally enforceable definition of hate speech. For the detection method, the authors present various settings of prompt tuning on a RoBERTa-large model. Prompt-tuning is a new parameter-efficient fine-tuning method that enables fine-tuning of large language models in low-resource settings, by freezing the model weights and updating a small set of parameters instead. ~\citeA{del2023respectful} evaluates zero-shot hate speech detection by simply prompting instruction-tuned models FLAN-T5~\citeA{chung2022scaling} and mT0~\citeA{muennighoff2022crosslingual}, and compare the performance with encoder-based language models such as the BERT family of models. They perform the evaluation on an extensive collection of 8 benchmark datasets containing online hate speech. Their results show that the instruction-tuned models have superior performance. 

Recently, the accessibility and ease of use of ChatGPT, along with its impressive performance has inspired a series of interesting exploratory efforts into using ChatGPT as a detector for many NLP tasks. Along this direction, authors in ~\citeA{huang2023chatgpt} have experimented with ChatGPT to understand how well it can detect implicit hate speech in Tweets, and also whether it can provide explanations for the reasoning. Their experiments demonstrate that ChatGPT has the potential to be used for such subjective tasks such as implicit hate speech detection. Furthermore, ChatGPT generated explanations also appear to have more clarity than human-written explanations, although there was no significant difference in informativeness. ChatGPT has also been evaluated for language-specific hate speech detection in Portuguese~\citeA{oliveira2023good} and results show that even without any fine-tuning, ChatGPT performs well in the detection task.

\section{Empirical Analysis}

In this section, we undertake several experiments utilizing representative LLMs to empirically assess their proficiency in identifying hate speech. Through these experiments, we address two primary research questions:

\begin{itemize}
\item \textbf{RQ1}: How robust are LLMs in classifying hate speech?
\item \textbf{RQ2}: How do various prompting techniques influence the hate speech detection efficacy of LLMs?
\end{itemize}

\subsection{Experiment Design}

In this subsection, we delve into the details of our experimental design, highlighting the critical decisions made to address the stated research questions. Paramount among these decisions were the choice of LLMs as the hate speech detector(classifier) and the dataset selection to rigorously assess the robustness of the chosen LLMs in detecting hate speech.

\subsubsection{LLM Selection}

As mentioned in Section \ref{sec:LLM}, numerous advanced LLMs are currently available, encompassing both open-source and proprietary options. From the open-source category, we opted for the Llama-2 model (7B parameters chat variant\footnote{https://huggingface.co/meta-llama/Llama-2-7b-chat-hf}) and the Falcon model (7B parameters chat variant\footnote{https://huggingface.co/tiiuae/falcon-7b-instruct}) based on their notable standings on the Open LLM Leaderboard hosted by Hugging Face\footnote{https://huggingface.co/spaces/HuggingFaceH4/open\_llm\_leaderboard}. For proprietary models, we selected GPT 3.5, which serves as a foundational benchmark for subsequent LLMs, including GPT 4. To enhance the reproducibility of our results, we utilized the specific version ``gpt-3.5-turbo-0613.'' The experiments involving the chosen LLMs—Llama-2, Falcon, and GPT 3.5—were conducted between September 25th and October 8th, 2023.

\subsubsection{Dataset Selection}

The primary objective of this study is to assess the robustness and reliability of selected LLM classifiers in identifying real-world hate speech. To achieve this, our evaluation criteria encompass: i) determining the classifier's capability to recognize general hate speech concepts; ii) gauging its sensitivity to hate speech directed at specific groups, such as women or Muslims; and iii) ascertaining any biases the classifier might exhibit towards extraneous features, like swear words or group identifiers.

\begin{figure}[h]
    \centering
    \subfloat[Label distribution]{%
        \includegraphics[width=0.3\linewidth]{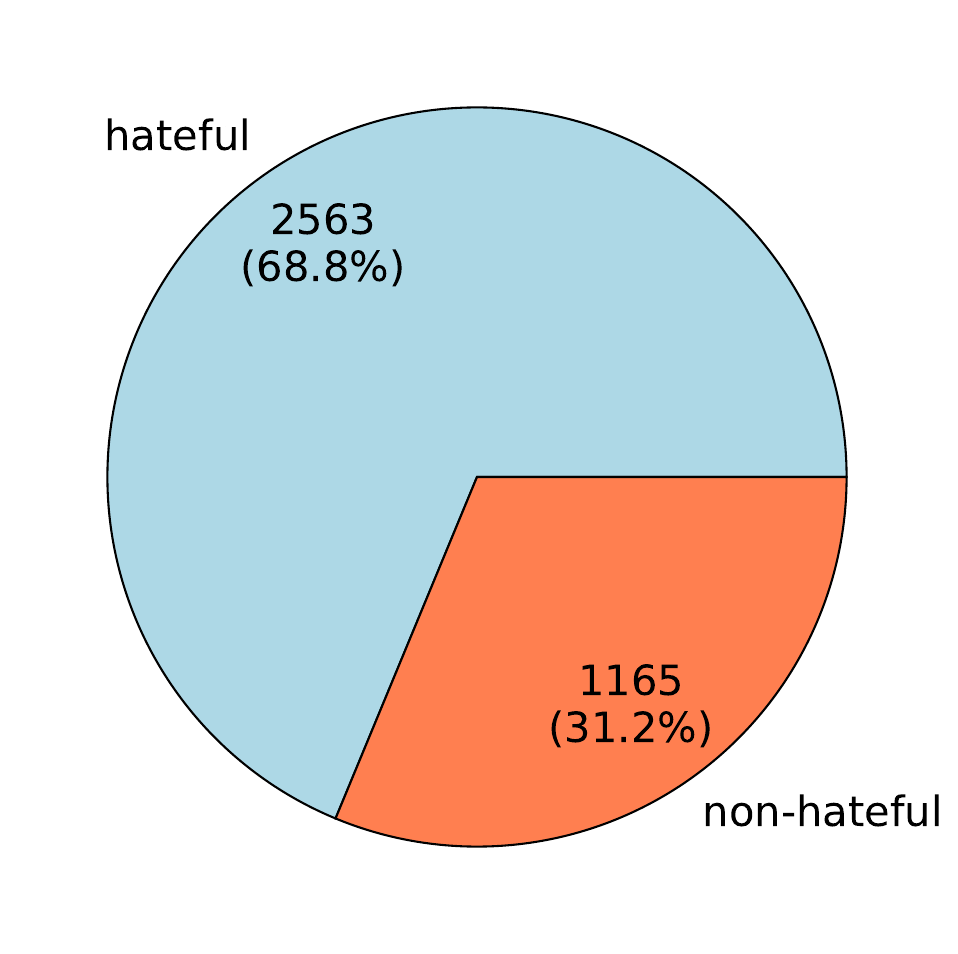}%
        \label{fig:label_dist}%
    }
    \subfloat[Directed vs general hate examples]{%
        \includegraphics[width=0.3\linewidth]{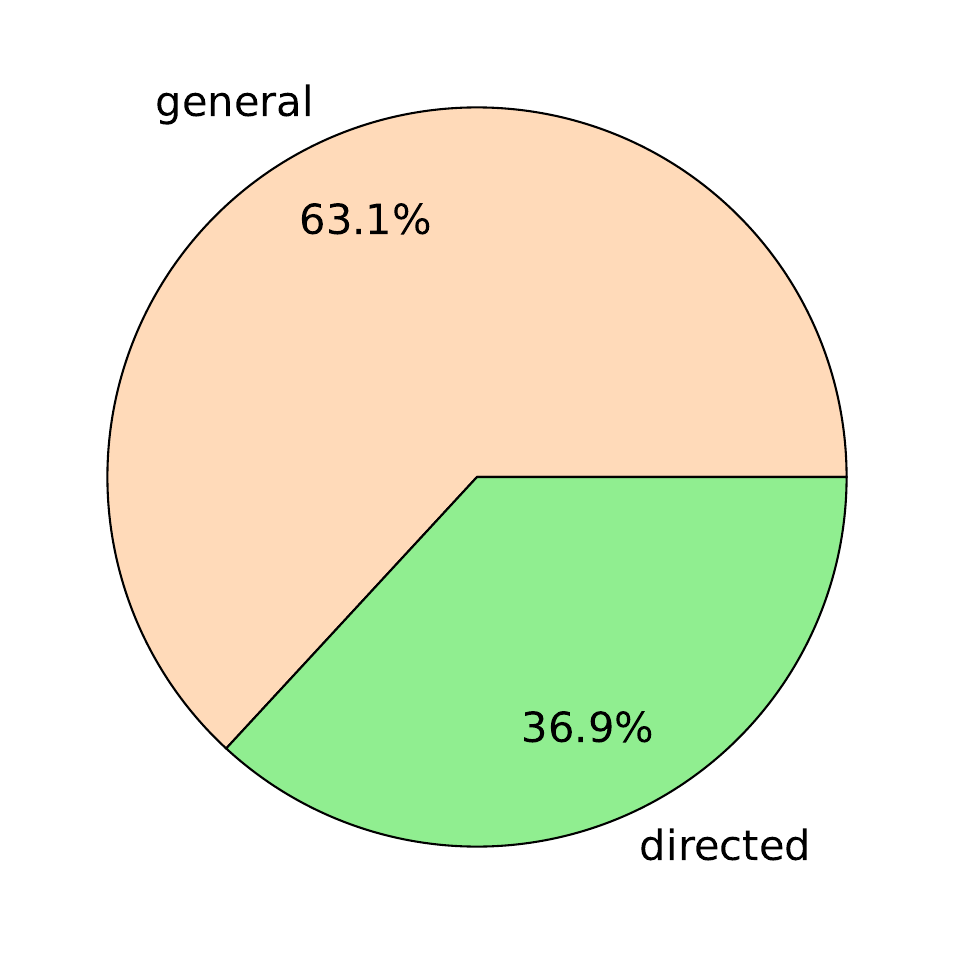}%
        \label{fig:hate_direction}%
    }
    \subfloat[Hate target distribution]{%
        \includegraphics[width=0.38\linewidth]{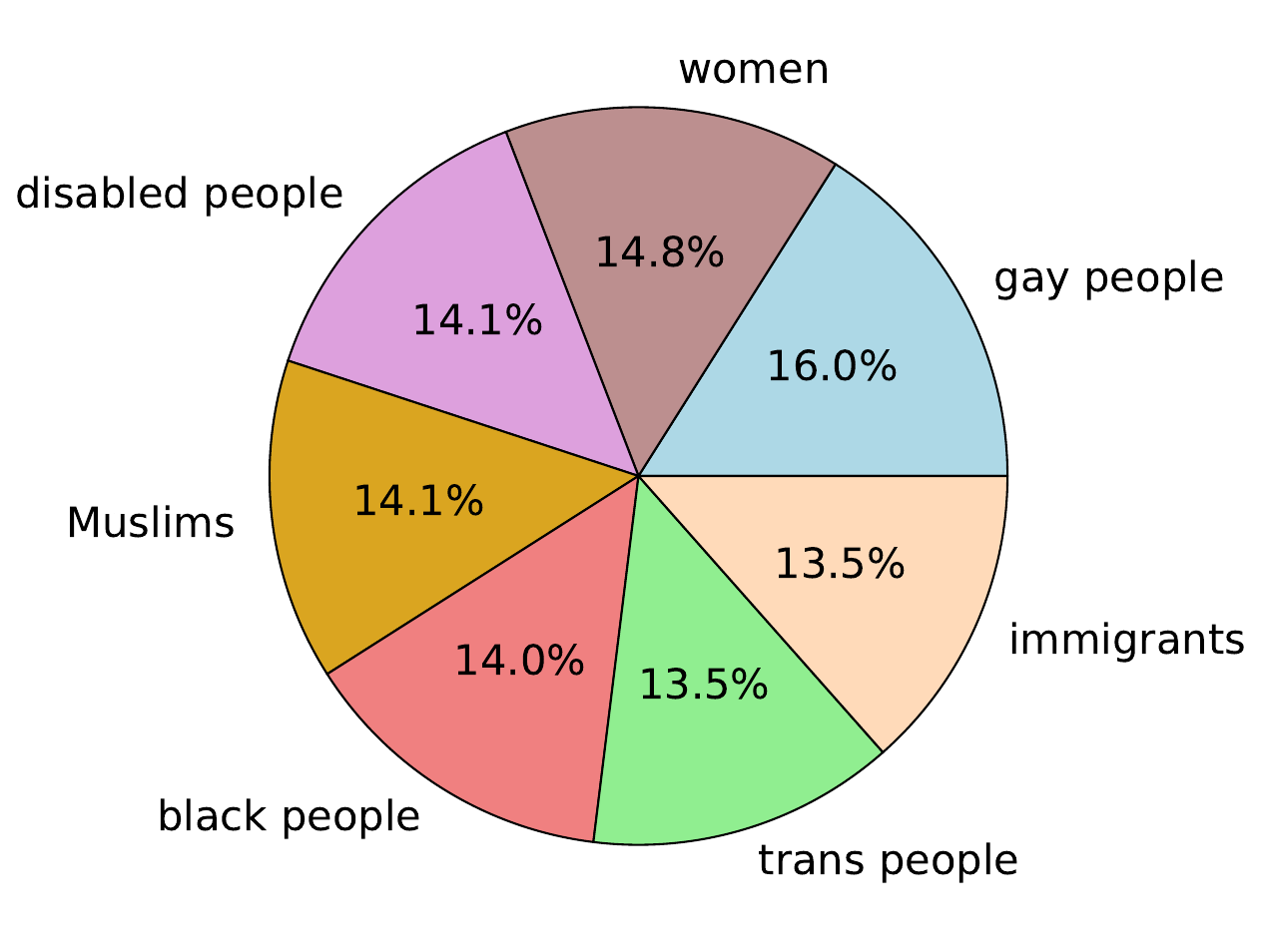}%
        \label{fig:hate_target}%
    }
    \caption{HateCheck dataset statistics}
    \label{fig:dataset}
\end{figure}

Consequently, we chose the HateCheck dataset~\citeA{rottger-etal-2021-hatecheck} as the primary hate speech dataset to evaluate the LLM classifiers. This dataset is meticulously designed to serve as a diagnostic tool for hate speech detection systems, aligning perfectly with our evaluation criteria. It features annotations that categorize hate speech as either 'directed' or 'general' and identifies the target of the hate. Furthermore, the non-hateful examples within this dataset are constructed to probe potential biases in hate speech classifiers towards spurious correlations. Consider the example: ``They are building a new dyke near the beach.'' The aim here is to discern if the classifier exhibits a bias towards the term ``dyke'' and misclassifies these examples as hate speech. For a comprehensive overview of the dataset statistics, refer to Figure \ref{fig:dataset}.

\subsubsection{Hate Speech Detection}

The concluding aspect of our experimental design pertains to the methodology detailing our hate speech classification approach using the chosen LLMs. We employed the LLMs in a zero-shot fashion (i.e., without specific training on hate speech datasets), presenting them with examples from the HateCheck dataset and instructing each LLM to determine the hatefulness of the input text. Consequently, it is imperative to discuss both the \textit{input prompt} utilized to query the LLM and the \textit{labeling method} adopted to categorize the LLM's response and decide the final classification label.\\

\noindent \textbf{Classification Prompt}

The input prompt is integral to LLM-based classifiers. Its primary objective is to concisely direct the LLM to analyze a provided text and determine the appropriate label, signifying whether the content is hateful or non-hateful. In our study, we incorporated the following prompt:

\begin{tcolorbox}[colback=blue!5!white,colframe=blue!75!black,title= Prompt]
  Is the following text hateful or not? Just answer in Yes or No. Don't provide explanations. \\

  text: \{hate\_speech\}
  
\end{tcolorbox}


\noindent \textbf{Labeling Method and Caveats}\\
The labeling method is employed to translate the text output of the LLM into binary class labels: 1 ('hate') and 0 ('non-hate'). When the LLM explicitly responds with 'Yes' or 'No', the label mapping process is straightforward. However, some scenarios necessitate a more nuanced approach to categorize the output:

\begin{itemize}
    \item Caveat 1: Deviation from Instructions: LLMs occasionally diverge from the provided directives and offer explanations alongside the label. In these instances, we manually reviewed the diverse, unique outputs, determining the appropriate labels grounded in keywords like 'Yes,' 'hateful,' 'No,' and 'not hateful.'
    \item Caveat 2: Activation of LLM Guardrails: Certain examples within the HateCheck dataset activate the LLM's built-in guardrails, designed to identify and mitigate hateful or offensive content processing. When these guardrails are triggered, the LLM yields a message indicating the presence of hate or offensive language, leading us to categorize such instances as hate speech.
\end{itemize}

\subsection{Experiment Results}

\subsubsection{RQ1: LLM's Hate Classification Performance}

Table \ref{tab:main_results} displays the efficacy of selected LLMs in classifying hate speech, using data from the HateCheck dataset. The performance metrics, derived from direct prompt outcomes, reveal that both GPT-3.5 and Llama 2 exhibit commendable efficiency, with accuracy and F1 scores ranging between 80-90\%. This underscores their proficiency in identifying hate speech. GPT-3.5 outperforms the others, an expected outcome given it has benefited from numerous advanced iterations of Reinforcement Learning from Human Feedback (RLHF) (from November 2022 onwards), and it contains more parameters than the other LLMs we considered. In contrast, Llama 2, despite its smaller 7B parameter model, delivers a performance that nearly matches GPT-3.5. The Falcon model, however, demonstrates inferior classification, performing below the level of random guessing. This disparity in performance between Llama 2 and Falcon can be attributed to the specific tuning conducted to optimize their pre-trained versions for chat compatibility. Another potential explanation is that the Llama 2 authors deliberately retained toxic data during pre-training to enhance downstream task generalization~\citeA{touvron2023llama}, positioning it as a more adept hate speech classifier than the Falcon model.\\


\begin{table}[b]
\centering
\begin{tabular}{@{}ccccccccc@{}}
\toprule
\textbf{LLM}     & \multicolumn{3}{c}{\textbf{Hate Class}}    & \multicolumn{3}{c}{\textbf{Non-Hate Class}} & \multicolumn{2}{c}{\textbf{Overall}} \\ \midrule
                 & \textbf{P}    & \textbf{R} & \textbf{F1}   & \textbf{P}  & \textbf{R}    & \textbf{F1}   & \textbf{Accuracy}  & \textbf{AUROC}  \\
\textbf{Falcon}  & 0.69 & 0.43          & 0.53 & 0.3           & 0.56 & 0.4  & 0.47 & 0.49 \\
\textbf{Llama 2} & 0.80 & \textbf{1.00} & 0.89 & \textbf{0.99} & 0.46 & 0.63 & 0.83 & 0.73 \\
\textbf{GPT 3.5} & \textbf{0.89} & 0.98       & \textbf{0.93} & 0.93        & \textbf{0.73} & \textbf{0.82} & \textbf{0.89}      & \textbf{0.85}   \\ \bottomrule
\end{tabular}%
\caption{Hate classification results: Precision(P), Recall(R), F1-score(F1) values are recorded for both ``Hate'' and ``Non-Hate'' classes. Highest performance under each metric is in \textbf{bold}.}
\label{tab:main_results}
\end{table}

\noindent \textbf{Error Analysis}\\
We conducted an error analysis to delve into the challenges the existing LLMs face in identifying hate speech and to pinpoint specific contexts where these models struggle to discern hate speech effectively. For this examination, we utilized the directionality annotations and target annotations from the HateCheck dataset. Within the realm of directionality, we assessed the proportion of misclassified hate speech samples, distinguishing between errors in identifying directed hate speech and those in discerning general hate speech. As shown in Table \ref{tab:main_results}, both Llama 2 and Falcon have equal error rates for directed and general hate speech, suggesting that these models possess comparable proficiency in detecting both types of hate speech. In contrast, GPT 3.5 exhibits a higher error rate for directed hate speech than general hate speech.
\begin{table}[t]
\resizebox{\columnwidth}{!}{%
\centering
\begin{tabular}{@{}cccccccccc@{}}
\toprule
\textbf{LLM}     & \multicolumn{2}{c}{\textbf{Directionality}} & \multicolumn{7}{c}{\textbf{Hate Target}}                       \\ \midrule
\textbf{} &
  \textbf{Directed} &
  \textbf{General} &
  \textbf{Women} &
  \textbf{Gay} &
  \textbf{Immigrants} &
  \textbf{Trans} &
  \textbf{Black} &
  \textbf{Muslims} &
  \textbf{Disabled} \\
\textbf{Falcon} &
  \textbf{53.7} &
  \textbf{59.0} &
  14.1 &
  13.0 &
  \textbf{14.3} &
  \textbf{13.1} &
  \textbf{15.4} &
  \textbf{15.6} &
  \textbf{14.6} \\
\textbf{Llama 2} & 0.2                  & 0.2                  & 9.7           & \textbf{15.6} & 5.4  & 5.9 & 12.0 & 8.2  & 9.0 \\
\textbf{GPT 3.5} & 0.6                  & 0.3                  & \textbf{47.6} & 7.9           & 14.2 & 6.3 & 3.2  & 14.2 & 6.3 \\ \bottomrule
\end{tabular}%
}
\caption{Error analysis: error rate (\%) under ``directionality'' and ``hate-target''. Highest error rate under each category is in \textbf{bold}.}
\label{tab:error_ana}
\end{table}
Subsequently, we assessed the error rates of the LLMs concerning different hate targets. The objective of this segment was to ascertain which target-associated hate speech poses the most significant detection challenges for the LLMs. As demonstrated in Table \ref{tab:error_ana}, the error rates for Llama 2 and Falcon regarding specific targets largely mirror the original distribution of these targets in the dataset. However, GPT 3.5 exhibits a disproportionately elevated error rate when identifying hate speech related to ``women.''\\

\noindent \textbf{Performance Attributed to Spurious Correlations Rather Than Proper Reasoning}

It is crucial to examine whether the notable classification performance of LLMs can be attributed to spurious correlations, such as categorizing a text as hate speech based solely on the presence of swear words or group identifiers, rather than substantive reasoning. This consideration is facilitated by the non-hate examples included in the HateCheck dataset, which contains elements like swear words and group identifiers used in non-hateful contexts. Evaluating the performance of LLMs in classifying these ``non-hate'' examples is essential to confirm their reliability as hate speech classifiers. As detailed in Table \ref{tab:main_results}, although Llama 2 demonstrates impressive classification accuracy for ``hate'' content, its performance diminishes in identifying non-hateful content, suggesting a reliance on spurious correlations. Conversely, GPT 3.5 maintains robust performance in classifying both ``hate'' and ``non-hate'' content.

\begin{figure}[b]
    \centering
    \includegraphics[width=0.6\linewidth]{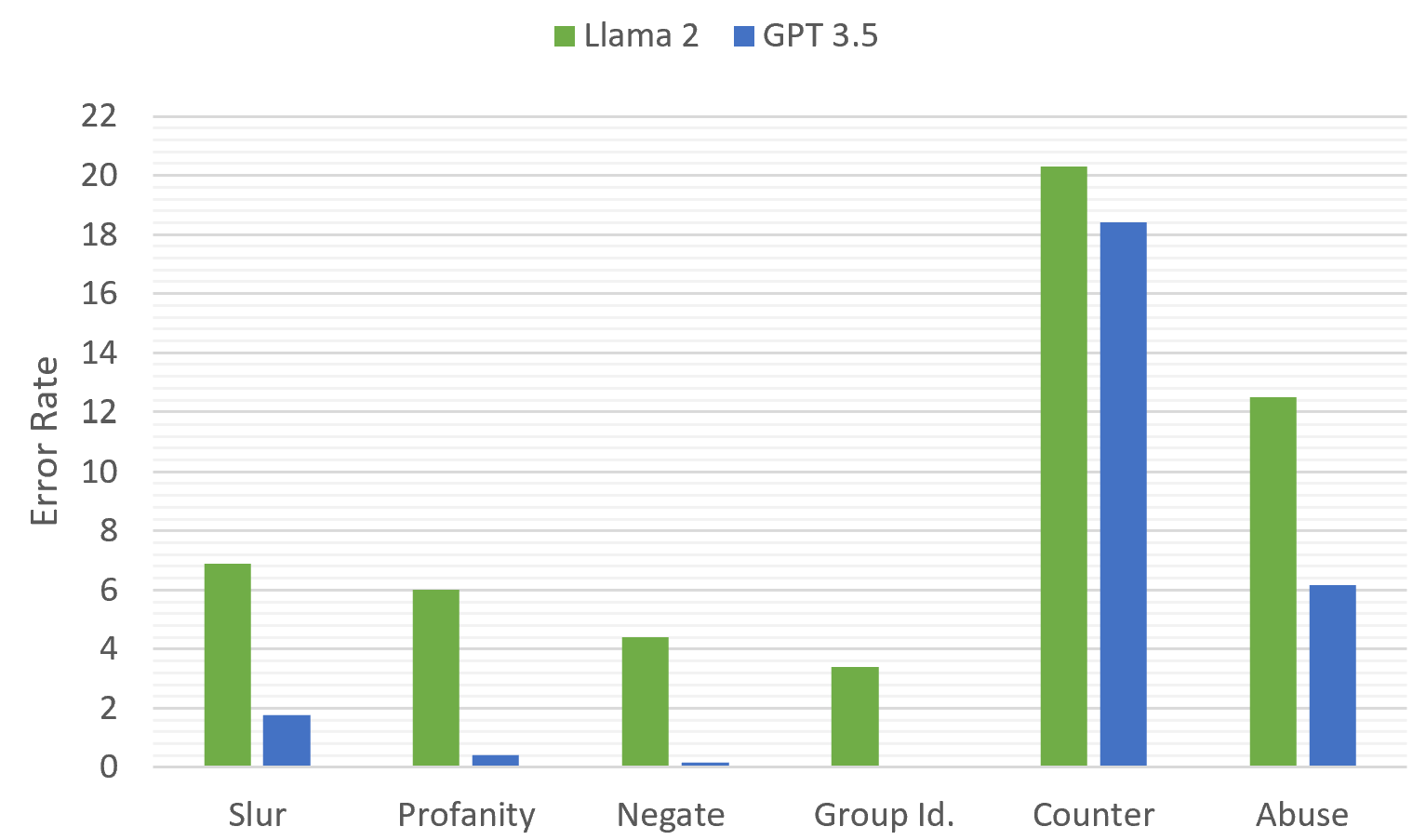}%
    \caption{Error analysis on non-hate class}
    \label{fig:spurious}
\end{figure}

We further investigated the specific types of spurious correlations influencing these LLMs using the functionality annotations of the HateCheck dataset. These annotations identify various categories of spurious correlations scenarios evident in non-hateful content, including ``slur'', ``profanity'', ``negate hateful statements'', ``group identifiers'', ``countering of hate speech through quoting or referencing hate speech examples'' and ``abuse targeted at objects, individuals, and non-protected groups.'' As illustrated in Figure \ref{fig:spurious}, Llama 2 exhibits more errors attributed to spurious correlations, further underlining its diminished performance in classifying the 'non-hate' category. Both Llama 2 and GPT 3.5 display heightened inaccuracies in distinguishing examples that counteract hate speech by referencing or quoting hate speech instances. This augmented error rate may be, in part, due to the labeling function, where specific counter-speech scenarios could trigger the LLM guardrails. As a result, the labeling function might mistakenly assume that the LLM's response to these examples implies a hate label. This underscores the significance of adequately addressing such scenarios when integrating LLMs into real-world hate speech detection frameworks.

\subsubsection{RQ2: Effect of Prompting}

\begin{figure}[]
    \centering
    \includegraphics[width=1.0\linewidth]{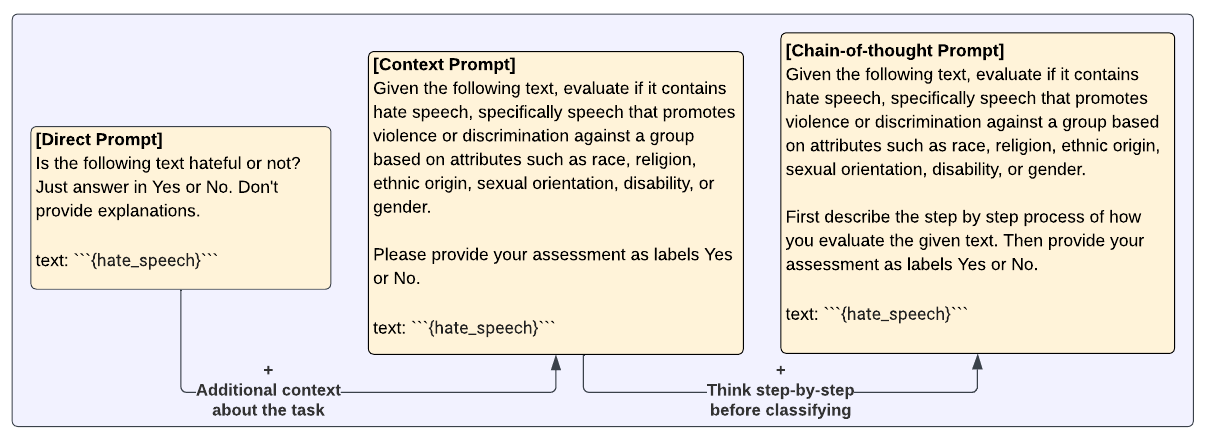}%
    \caption{Prompt templates used for hate speech classification}
    \label{fig:prompt}
\end{figure}

The input prompt plays an indispensable role in LLM-based classifiers. Generally, the efficacy of an LLM in classifying text is intrinsically tied to the quality of the input prompt. In light of this, we conducted an extended experiment involving the top-performing LLM, GPT 3.5, to explore the impact of various prompts on classification performance. As illustrated in Figure \ref{fig:prompt}, we introduced two additional prompt types, referred to as \textit{context prompt}, and \textit{chain-of-thought(COT) prompt}. 

Table \ref{tab:prompt_results} presents the classification results of GPT 3.5 using different prompts employed in our study. Intuitively, we anticipated the performance of the LLM classifier to improve as we transitioned through the prompts from left to right in Figure \ref{fig:prompt}, particularly given the additional context and incorporation of the COT method. However, unexpectedly, the direct concise prompt yielded the most superior performance out of the three prompts. One potential rationale for this result is that an overly complex prompt, paired with the inherently intricate nature of hate speech detection, might obscure the LLM's understanding of the task rather than clarifying it. Another explanation aligns with recent findings on LLMs, suggesting that performance peaks when vital information is positioned at the beginning or end of the input context and diminishes substantially when models must retrieve relevant information from the middle of lengthy contexts~\citeA{liu2023lost}.

\begin{table}[b]
\centering
\begin{tabular}{@{}ccccccccc@{}}
\toprule
\textbf{Prompt}     & \multicolumn{3}{c}{\textbf{Hate Class}}    & \multicolumn{3}{c}{\textbf{Non-Hate Class}} & \multicolumn{2}{c}{\textbf{Overall}} \\ \midrule
                 & \textbf{P}    & \textbf{R} & \textbf{F1}   & \textbf{P}  & \textbf{R}    & \textbf{F1}   & \textbf{Accuracy}  & \textbf{AUROC}  \\
\textbf{Direct}  & 0.89 & \textbf{0.98}       & \textbf{0.93} & \textbf{0.93}        & 0.73 & \textbf{0.82} & \textbf{0.89}      & \textbf{0.85} \\
\textbf{Context} & \textbf{0.91} & 0.85 & 0.88 & 0.71 & \textbf{0.82} & 0.76 & 0.84 & 0.83 \\
\textbf{COT} & 0.88 & 0.81 & 0.84 & 0.69 & 0.79 & 0.74 & 0.80 & 0.79   \\ \bottomrule
\end{tabular}%
\caption{GPT 3.5's hate classification results with different prompts: Precision(P), Recall(R), F1-score(F1) values are recorded for both ``Hate'' and ``Non-Hate'' classes. Highest performance under each metric is in \textbf{bold}.}
\label{tab:prompt_results}
\end{table}

\subsection{Discussion}
In addressing the two research questions posed, our findings offer significant insights into the robustness and nuances of LLMs in hate speech classification.

\subsubsection{Answering RQ1: LLM's Robustness in Classifying Hate Speech}
For \textbf{RQ1}, the GPT-3.5 and Llama 2 models proved their robustness in classifying hate speech, boasting accuracy and F1 scores between 80-90\%. Despite its fewer parameters, Llama 2 nearly matches the performance of GPT-3.5, although GPT-3.5 remains superior. We attribute this to its advanced RLHF iterations and larger parameter size. Falcon, conversely, demonstrated subpar performance, indicating its unsuitability for reliable hate speech classification. The error analysis further enriched our understanding. While Llama 2 and Falcon demonstrated equal proficiency in detecting directed and general hate speech, GPT-3.5 showed a higher error rate for directed hate speech. Additionally, it exhibited an increased error rate in identifying hate speech targeted at women, indicating potential areas for improvement in its training and calibration. Llama 2's diminished performance in classifying non-hateful content hinted at its reliance on spurious correlations. Both Llama 2 and GPT-3.5 were challenged in scenarios involving the counteraction of hate speech through referencing or quoting hateful content, pinpointing a need to refine the LLMs' handling of such contexts.

\subsubsection{Answering RQ2: Influence of Prompting Techniques}
As for \textbf{RQ2}, the efficacy of LLMs is notably influenced by the employed prompting techniques. Contrary to our anticipation that more complex prompts (such as context and chain-of-thought prompts) would enhance classification performance, the direct concise prompts delivered best results. It suggests that simplicity and conciseness in prompts might facilitate clearer hate speech detection task comprehension for LLMs, leading to more accurate classifications.

\section{Best Practices and Pro Tips}

\subsubsection*{Optimizing LLM Performance}
When utilizing LLMs as hate speech classifiers, certain practices can optimize their performance and reliability.

\begin{itemize}
    \item \textbf{Select Appropriate LLMs}: GPT-3.5 and Llama 2 have shown notable efficacy; however, it’s crucial to consider the specific needs and contexts of the application. Evaluate multiple models to identify which offers the best balance of accuracy and computational efficiency.
    \item \textbf{Input Prompt}: Direct and concise prompts have been shown to be more effective. Avoid overly complex prompts that could potentially confuse the model or dilute the task’s clarity. Experiment with various prompt designs to identify which yields optimal performance for the specific LLM and classification task.
    \item \textbf{Error Analysis}: Conduct detailed error analyses to identify specific areas where the LLM struggles, and consider this information when fine-tuning or selecting models for deployment.
    \item \textbf{Labeling Function}: The labeling function plays a pivotal role in the performance of LLMs in classification tasks. It’s essential to optimize and test various labeling functions to ensure that they are accurate and reliable, avoiding misclassifications especially in complex scenarios like counter-speech.
\end{itemize}

\subsubsection*{Mitigating the Influence of Spurious Correlations}
The risk of LLMs relying on spurious correlations, as observed with Llama 2, underscores the necessity of specific strategies to mitigate such influences.

\begin{itemize}
    \item \textbf{Balanced Fine-tuning}: Conduct additional fine-tuning of the LLM with balanced training data that includes diverse examples of hate speech and non-hate speech, reducing the model’s reliance on specific words or phrases as indicators of hate speech.
    \item \textbf{Functionality Annotations}: Leverage functionality annotations to identify and analyze potential spurious correlations, enabling the refinement of the model’s classification capabilities.
    \item \textbf{Real-world Testing}: Test the LLMs in real-world scenarios to assess their performance beyond controlled experiments. Adapt and refine the models continuously based on the emerging data and classification challenges.
\end{itemize}

Incorporating these insights and practices will be instrumental in enhancing the reliability, accuracy, and fairness of LLMs in hate speech classification, ensuring they are a valuable tool in combating online hate while preserving freedom of expression.

\section{Conclusion}

In our study, we provided a detailed look into the progression of language models for hate speech classification, from the days of pre-LLMs to the modern era of sophisticated LLMs like GPT. Earlier language models, often needed significant fine-tuning to work well, but new LLMs, like GPT-3.5 and Llama 2, have shown they can be  effective at identifying some forms of hate speech right out of the box, even in zero and few shot settings. 

We explored the capabilities of three LLMs, GPT-3.5, Llama 2 and Falcon, on the HateCheck dataset to gain deeper insights into their abilities and challenges in classifying hate speech. From our experiments, a few key points stood out: GPT-3.5 and Llama-2 were quite effective overall with accuracy levels between 80-90\%, but Falcon lagged behind considerably. As we discussed, this may be an artifact of what data was used to train Falcon. When we looked into the nuances of hate speech, like understanding who the hate was directed at, all of these models faced challenges and their performance declined considerably. For instance, GPT 3.5 struggled particularly with recognizing hate directed towards women. We also found through experimentation that clear and straightforward prompts worked best, hinting that simplicity of classification instructions may be key for effective classification performance.

Hate speech classification remains a challenging area for many reasons, not just due to its nuanced nature but also the ethical concerns around data collection and especially labeling. LLMs, even in zero and few shot settings, present a potential exciting way forward. While they are promising, there is still much to understand and refine. We hope our findings and recommendations from this study offer a useful guide for those looking to delve further into the capabilities of LLMs for managing online hate. Forging towards a safer, more inclusive digital landscape for everyone.

\bibliographystyle{apacite}
\bibliography{chapter}

\begin{thebibliography}{}

\bibitem [\protect \citeauthoryear {%
Arslan%
\ \protect \BOthers {.}}{%
Arslan%
\ \protect \BOthers {.}}{%
{\protect \APACyear {2021}}%
}]{%
arslan2021comparison}
\APACinsertmetastar {%
arslan2021comparison}%
\begin{APACrefauthors}%
Arslan, Y.%
\BCBT {}\ \BOthersPeriod {.}
\end{APACrefauthors}%
\unskip\
\newblock
\APACrefYearMonthDay{2021}{}{}.
\newblock
{\BBOQ}\APACrefatitle {A comparison of pre-trained language models for
  multi-class text classification in the financial domain} {A comparison of
  pre-trained language models for multi-class text classification in the
  financial domain}.{\BBCQ}
\newblock
\BIn{} \APACrefbtitle {Companion Proceedings of the Web Conference 2021}
  {Companion proceedings of the web conference 2021}\ (\BPGS\ 260--268).
\PrintBackRefs{\CurrentBib}

\bibitem [\protect \citeauthoryear {%
Bhattacharjee%
\ \BBA {} Liu%
}{%
Bhattacharjee%
\ \BBA {} Liu%
}{%
{\protect \APACyear {2023}}%
}]{%
bhattacharjee2023fighting}
\APACinsertmetastar {%
bhattacharjee2023fighting}%
\begin{APACrefauthors}%
Bhattacharjee, A.%
\BCBT {}\ \BBA {} Liu, H.%
\end{APACrefauthors}%
\unskip\
\newblock
\APACrefYearMonthDay{2023}{}{}.
\newblock
{\BBOQ}\APACrefatitle {Fighting Fire with Fire: Can ChatGPT Detect AI-generated
  Text?} {Fighting fire with fire: Can chatgpt detect ai-generated
  text?}{\BBCQ}
\newblock
\APACjournalVolNumPages{arXiv preprint arXiv:2308.01284}{}{}{}.
\PrintBackRefs{\CurrentBib}

\bibitem [\protect \citeauthoryear {%
Brown%
\ \protect \BOthers {.}}{%
Brown%
\ \protect \BOthers {.}}{%
{\protect \APACyear {2020}}%
}]{%
brown2020language}
\APACinsertmetastar {%
brown2020language}%
\begin{APACrefauthors}%
Brown, T.%
\BCBT {}\ \BOthersPeriod {.}
\end{APACrefauthors}%
\unskip\
\newblock
\APACrefYearMonthDay{2020}{}{}.
\newblock
{\BBOQ}\APACrefatitle {Language models are few-shot learners} {Language models
  are few-shot learners}.{\BBCQ}
\newblock
\APACjournalVolNumPages{Advances in neural information processing
  systems}{33}{}{1877--1901}.
\PrintBackRefs{\CurrentBib}

\bibitem [\protect \citeauthoryear {%
Chiu%
, Collins%
\BCBL {}\ \BBA {} Alexander%
}{%
Chiu%
\ \protect \BOthers {.}}{%
{\protect \APACyear {2021}}%
}]{%
chiu2021detecting}
\APACinsertmetastar {%
chiu2021detecting}%
\begin{APACrefauthors}%
Chiu, K\BHBI L.%
, Collins, A.%
\BCBL {}\ \BBA {} Alexander, R.%
\end{APACrefauthors}%
\unskip\
\newblock
\APACrefYearMonthDay{2021}{}{}.
\newblock
{\BBOQ}\APACrefatitle {Detecting hate speech with gpt-3} {Detecting hate speech
  with gpt-3}.{\BBCQ}
\newblock
\APACjournalVolNumPages{arXiv preprint arXiv:2103.12407}{}{}{}.
\PrintBackRefs{\CurrentBib}

\bibitem [\protect \citeauthoryear {%
Chung%
\ \protect \BOthers {.}}{%
Chung%
\ \protect \BOthers {.}}{%
{\protect \APACyear {2022}}%
}]{%
chung2022scaling}
\APACinsertmetastar {%
chung2022scaling}%
\begin{APACrefauthors}%
Chung, H\BPBI W.%
\BCBT {}\ \BOthersPeriod {.}
\end{APACrefauthors}%
\unskip\
\newblock
\APACrefYearMonthDay{2022}{}{}.
\newblock
{\BBOQ}\APACrefatitle {Scaling instruction-finetuned language models} {Scaling
  instruction-finetuned language models}.{\BBCQ}
\newblock
\APACjournalVolNumPages{arXiv preprint arXiv:2210.11416}{}{}{}.
\PrintBackRefs{\CurrentBib}

\bibitem [\protect \citeauthoryear {%
Czapla%
\ \protect \BOthers {.}}{%
Czapla%
\ \protect \BOthers {.}}{%
{\protect \APACyear {2019}}%
}]{%
czapla2019universal}
\APACinsertmetastar {%
czapla2019universal}%
\begin{APACrefauthors}%
Czapla, P.%
\BCBT {}\ \BOthersPeriod {.}
\end{APACrefauthors}%
\unskip\
\newblock
\APACrefYearMonthDay{2019}{}{}.
\newblock
{\BBOQ}\APACrefatitle {Universal language model fine-tuning for polish hate
  speech detection} {Universal language model fine-tuning for polish hate
  speech detection}.{\BBCQ}
\newblock
\APACjournalVolNumPages{Proceedings ofthePolEval2019Workshop}{}{}{149}.
\PrintBackRefs{\CurrentBib}

\bibitem [\protect \citeauthoryear {%
Dat%
\ \BBA {} Tuan%
}{%
Dat%
\ \BBA {} Tuan%
}{%
{\protect \APACyear {2020}}%
}]{%
dat2020phobert}
\APACinsertmetastar {%
dat2020phobert}%
\begin{APACrefauthors}%
Dat, N.%
\BCBT {}\ \BBA {} Tuan, N.%
\end{APACrefauthors}%
\unskip\
\newblock
\APACrefYearMonthDay{2020}{}{}.
\newblock
{\BBOQ}\APACrefatitle {Phobert: Pre-trained language models for Vietnamese}
  {Phobert: Pre-trained language models for vietnamese}.{\BBCQ}
\newblock
\APACjournalVolNumPages{Findings of the Association for Computational
  Linguistics: EMNLP}{2020}{}{1037--1042}.
\PrintBackRefs{\CurrentBib}

\bibitem [\protect \citeauthoryear {%
Del~Arco%
, Nozza%
\BCBL {}\ \BBA {} Hovy%
}{%
Del~Arco%
\ \protect \BOthers {.}}{%
{\protect \APACyear {2023}}%
}]{%
del2023respectful}
\APACinsertmetastar {%
del2023respectful}%
\begin{APACrefauthors}%
Del~Arco, F\BPBI M\BPBI P.%
, Nozza, D.%
\BCBL {}\ \BBA {} Hovy, D.%
\end{APACrefauthors}%
\unskip\
\newblock
\APACrefYearMonthDay{2023}{}{}.
\newblock
{\BBOQ}\APACrefatitle {Respectful or Toxic? Using Zero-Shot Learning with
  Language Models to Detect Hate Speech} {Respectful or toxic? using zero-shot
  learning with language models to detect hate speech}.{\BBCQ}
\newblock
\BIn{} \APACrefbtitle {The 7th Workshop on Online Abuse and Harms (WOAH)} {The
  7th workshop on online abuse and harms (woah)}\ (\BPGS\ 60--68).
\PrintBackRefs{\CurrentBib}

\bibitem [\protect \citeauthoryear {%
Del~Vigna12%
\ \protect \BOthers {.}}{%
Del~Vigna12%
\ \protect \BOthers {.}}{%
{\protect \APACyear {2017}}%
}]{%
del2017hate}
\APACinsertmetastar {%
del2017hate}%
\begin{APACrefauthors}%
Del~Vigna12, F.%
\BCBT {}\ \BOthersPeriod {.}
\end{APACrefauthors}%
\unskip\
\newblock
\APACrefYearMonthDay{2017}{}{}.
\newblock
{\BBOQ}\APACrefatitle {Hate me, hate me not: Hate speech detection on facebook}
  {Hate me, hate me not: Hate speech detection on facebook}.{\BBCQ}
\newblock
\BIn{} \APACrefbtitle {Proceedings of the first Italian conference on
  cybersecurity (ITASEC17)} {Proceedings of the first italian conference on
  cybersecurity (itasec17)}\ (\BPGS\ 86--95).
\PrintBackRefs{\CurrentBib}

\bibitem [\protect \citeauthoryear {%
Devlin%
\ \protect \BOthers {.}}{%
Devlin%
\ \protect \BOthers {.}}{%
{\protect \APACyear {2018}}%
}]{%
devlin2018bert}
\APACinsertmetastar {%
devlin2018bert}%
\begin{APACrefauthors}%
Devlin, J.%
\BCBT {}\ \BOthersPeriod {.}
\end{APACrefauthors}%
\unskip\
\newblock
\APACrefYearMonthDay{2018}{}{}.
\newblock
{\BBOQ}\APACrefatitle {Bert: Pre-training of deep bidirectional transformers
  for language understanding} {Bert: Pre-training of deep bidirectional
  transformers for language understanding}.{\BBCQ}
\newblock
\APACjournalVolNumPages{arXiv preprint arXiv:1810.04805}{}{}{}.
\PrintBackRefs{\CurrentBib}

\bibitem [\protect \citeauthoryear {%
Garland%
\ \protect \BOthers {.}}{%
Garland%
\ \protect \BOthers {.}}{%
{\protect \APACyear {2020}}%
}]{%
garlandCounterClassifier}
\APACinsertmetastar {%
garlandCounterClassifier}%
\begin{APACrefauthors}%
Garland, J.%
\BCBT {}\ \BOthersPeriod {.}
\end{APACrefauthors}%
\unskip\
\newblock
\APACrefYearMonthDay{2020}{{\APACmonth{11}}}{}.
\newblock
{\BBOQ}\APACrefatitle {Countering hate on social media: Large scale
  classification of hate and counter speech} {Countering hate on social media:
  Large scale classification of hate and counter speech}.{\BBCQ}
\newblock
\BIn{} \APACrefbtitle {Proceedings of the Fourth Workshop on Online Abuse and
  Harms} {Proceedings of the fourth workshop on online abuse and harms}\
  (\BPGS\ 102--112).
\newblock
\APACaddressPublisher{}{Association for Computational Linguistics}.
\PrintBackRefs{\CurrentBib}

\bibitem [\protect \citeauthoryear {%
Garland%
\ \protect \BOthers {.}}{%
Garland%
\ \protect \BOthers {.}}{%
{\protect \APACyear {2022}}%
}]{%
garland2020impact}
\APACinsertmetastar {%
garland2020impact}%
\begin{APACrefauthors}%
Garland, J.%
\BCBT {}\ \BOthersPeriod {.}
\end{APACrefauthors}%
\unskip\
\newblock
\APACrefYearMonthDay{2022}{}{}.
\newblock
{\BBOQ}\APACrefatitle {Impact and dynamics of hate and counter speech online}
  {Impact and dynamics of hate and counter speech online}.{\BBCQ}
\newblock
\APACjournalVolNumPages{EPJ Data Science}{11}{1}{3}.
\PrintBackRefs{\CurrentBib}

\bibitem [\protect \citeauthoryear {%
Gilardi%
, Alizadeh%
\BCBL {}\ \BBA {} Kubli%
}{%
Gilardi%
\ \protect \BOthers {.}}{%
{\protect \APACyear {2023}}%
}]{%
gilardi2023chatgpt}
\APACinsertmetastar {%
gilardi2023chatgpt}%
\begin{APACrefauthors}%
Gilardi, F.%
, Alizadeh, M.%
\BCBL {}\ \BBA {} Kubli, M.%
\end{APACrefauthors}%
\unskip\
\newblock
\APACrefYearMonthDay{2023}{}{}.
\newblock
{\BBOQ}\APACrefatitle {Chatgpt outperforms crowd-workers for text-annotation
  tasks} {Chatgpt outperforms crowd-workers for text-annotation tasks}.{\BBCQ}
\newblock
\APACjournalVolNumPages{arXiv preprint arXiv:2303.15056}{}{}{}.
\PrintBackRefs{\CurrentBib}

\bibitem [\protect \citeauthoryear {%
Han%
\ \BBA {} Tang%
}{%
Han%
\ \BBA {} Tang%
}{%
{\protect \APACyear {2022}}%
}]{%
han2022designing}
\APACinsertmetastar {%
han2022designing}%
\begin{APACrefauthors}%
Han, L.%
\BCBT {}\ \BBA {} Tang, H.%
\end{APACrefauthors}%
\unskip\
\newblock
\APACrefYearMonthDay{2022}{}{}.
\newblock
{\BBOQ}\APACrefatitle {Designing of Prompts for Hate Speech Recognition with
  In-Context Learning} {Designing of prompts for hate speech recognition with
  in-context learning}.{\BBCQ}
\newblock
\BIn{} \APACrefbtitle {2022 International Conference on Computational Science
  and Computational Intelligence (CSCI)} {2022 international conference on
  computational science and computational intelligence (csci)}\ (\BPGS\
  319--320).
\PrintBackRefs{\CurrentBib}

\bibitem [\protect \citeauthoryear {%
Howard%
\ \BBA {} Ruder%
}{%
Howard%
\ \BBA {} Ruder%
}{%
{\protect \APACyear {2018}}%
}]{%
howard2018universal}
\APACinsertmetastar {%
howard2018universal}%
\begin{APACrefauthors}%
Howard, J.%
\BCBT {}\ \BBA {} Ruder, S.%
\end{APACrefauthors}%
\unskip\
\newblock
\APACrefYearMonthDay{2018}{}{}.
\newblock
{\BBOQ}\APACrefatitle {Universal language model fine-tuning for text
  classification} {Universal language model fine-tuning for text
  classification}.{\BBCQ}
\newblock
\APACjournalVolNumPages{arXiv preprint arXiv:1801.06146}{}{}{}.
\PrintBackRefs{\CurrentBib}

\bibitem [\protect \citeauthoryear {%
Huang%
, Kwak%
\BCBL {}\ \BBA {} An%
}{%
Huang%
\ \protect \BOthers {.}}{%
{\protect \APACyear {2023}}%
}]{%
huang2023chatgpt}
\APACinsertmetastar {%
huang2023chatgpt}%
\begin{APACrefauthors}%
Huang, F.%
, Kwak, H.%
\BCBL {}\ \BBA {} An, J.%
\end{APACrefauthors}%
\unskip\
\newblock
\APACrefYearMonthDay{2023}{}{}.
\newblock
{\BBOQ}\APACrefatitle {Is chatgpt better than human annotators? potential and
  limitations of chatgpt in explaining implicit hate speech} {Is chatgpt better
  than human annotators? potential and limitations of chatgpt in explaining
  implicit hate speech}.{\BBCQ}
\newblock
\APACjournalVolNumPages{arXiv preprint arXiv:2302.07736}{}{}{}.
\PrintBackRefs{\CurrentBib}

\bibitem [\protect \citeauthoryear {%
Kant%
\ \protect \BOthers {.}}{%
Kant%
\ \protect \BOthers {.}}{%
{\protect \APACyear {2018}}%
}]{%
kant2018practical}
\APACinsertmetastar {%
kant2018practical}%
\begin{APACrefauthors}%
Kant, N.%
\BCBT {}\ \BOthersPeriod {.}
\end{APACrefauthors}%
\unskip\
\newblock
\APACrefYearMonthDay{2018}{}{}.
\newblock
{\BBOQ}\APACrefatitle {Practical text classification with large pre-trained
  language models} {Practical text classification with large pre-trained
  language models}.{\BBCQ}
\newblock
\APACjournalVolNumPages{arXiv preprint arXiv:1812.01207}{}{}{}.
\PrintBackRefs{\CurrentBib}

\bibitem [\protect \citeauthoryear {%
Lample%
\ \BBA {} Conneau%
}{%
Lample%
\ \BBA {} Conneau%
}{%
{\protect \APACyear {2019}}%
}]{%
lample2019cross}
\APACinsertmetastar {%
lample2019cross}%
\begin{APACrefauthors}%
Lample, G.%
\BCBT {}\ \BBA {} Conneau, A.%
\end{APACrefauthors}%
\unskip\
\newblock
\APACrefYearMonthDay{2019}{}{}.
\newblock
{\BBOQ}\APACrefatitle {Cross-lingual language model pretraining} {Cross-lingual
  language model pretraining}.{\BBCQ}
\newblock
\APACjournalVolNumPages{arXiv preprint arXiv:1901.07291}{}{}{}.
\PrintBackRefs{\CurrentBib}

\bibitem [\protect \citeauthoryear {%
Lavergne%
\ \protect \BOthers {.}}{%
Lavergne%
\ \protect \BOthers {.}}{%
{\protect \APACyear {2020}}%
}]{%
lavergne2020thenorth}
\APACinsertmetastar {%
lavergne2020thenorth}%
\begin{APACrefauthors}%
Lavergne, E.%
\BCBT {}\ \BOthersPeriod {.}
\end{APACrefauthors}%
\unskip\
\newblock
\APACrefYearMonthDay{2020}{}{}.
\newblock
{\BBOQ}\APACrefatitle {Thenorth@ haspeede 2: Bert-based language model
  fine-tuning for italian hate speech detection} {Thenorth@ haspeede 2:
  Bert-based language model fine-tuning for italian hate speech
  detection}.{\BBCQ}
\newblock
\BIn{} \APACrefbtitle {7th Evaluation Campaign of Natural Language Processing
  and Speech Tools for Italian. Final Workshop, EVALITA} {7th evaluation
  campaign of natural language processing and speech tools for italian. final
  workshop, evalita}\ (\BVOL\ 2765).
\PrintBackRefs{\CurrentBib}

\bibitem [\protect \citeauthoryear {%
Lewis%
\ \protect \BOthers {.}}{%
Lewis%
\ \protect \BOthers {.}}{%
{\protect \APACyear {2019}}%
}]{%
lewis2019bart}
\APACinsertmetastar {%
lewis2019bart}%
\begin{APACrefauthors}%
Lewis, M.%
\BCBT {}\ \BOthersPeriod {.}
\end{APACrefauthors}%
\unskip\
\newblock
\APACrefYearMonthDay{2019}{}{}.
\newblock
{\BBOQ}\APACrefatitle {Bart: Denoising sequence-to-sequence pre-training for
  natural language generation, translation, and comprehension} {Bart: Denoising
  sequence-to-sequence pre-training for natural language generation,
  translation, and comprehension}.{\BBCQ}
\newblock
\APACjournalVolNumPages{arXiv preprint arXiv:1910.13461}{}{}{}.
\PrintBackRefs{\CurrentBib}

\bibitem [\protect \citeauthoryear {%
N\BPBI F.~Liu%
\ \protect \BOthers {.}}{%
N\BPBI F.~Liu%
\ \protect \BOthers {.}}{%
{\protect \APACyear {2023}}%
}]{%
liu2023lost}
\APACinsertmetastar {%
liu2023lost}%
\begin{APACrefauthors}%
Liu, N\BPBI F.%
\BCBT {}\ \BOthersPeriod {.}
\end{APACrefauthors}%
\unskip\
\newblock
\APACrefYearMonthDay{2023}{}{}.
\newblock
{\BBOQ}\APACrefatitle {Lost in the middle: How language models use long
  contexts} {Lost in the middle: How language models use long contexts}.{\BBCQ}
\newblock
\APACjournalVolNumPages{arXiv preprint arXiv:2307.03172}{}{}{}.
\PrintBackRefs{\CurrentBib}

\bibitem [\protect \citeauthoryear {%
Y.~Liu%
\ \protect \BOthers {.}}{%
Y.~Liu%
\ \protect \BOthers {.}}{%
{\protect \APACyear {2019}}%
}]{%
liu2019roberta}
\APACinsertmetastar {%
liu2019roberta}%
\begin{APACrefauthors}%
Liu, Y.%
\BCBT {}\ \BOthersPeriod {.}
\end{APACrefauthors}%
\unskip\
\newblock
\APACrefYearMonthDay{2019}{}{}.
\newblock
{\BBOQ}\APACrefatitle {Roberta: A robustly optimized bert pretraining approach}
  {Roberta: A robustly optimized bert pretraining approach}.{\BBCQ}
\newblock
\APACjournalVolNumPages{arXiv preprint arXiv:1907.11692}{}{}{}.
\PrintBackRefs{\CurrentBib}

\bibitem [\protect \citeauthoryear {%
Luo%
\ \protect \BOthers {.}}{%
Luo%
\ \protect \BOthers {.}}{%
{\protect \APACyear {2023}}%
}]{%
luo2023towards}
\APACinsertmetastar {%
luo2023towards}%
\begin{APACrefauthors}%
Luo, C\BPBI F.%
\BCBT {}\ \BOthersPeriod {.}
\end{APACrefauthors}%
\unskip\
\newblock
\APACrefYearMonthDay{2023}{}{}.
\newblock
{\BBOQ}\APACrefatitle {Towards Legally Enforceable Hate Speech Detection for
  Public Forums} {Towards legally enforceable hate speech detection for public
  forums}.{\BBCQ}
\newblock
\APACjournalVolNumPages{arXiv preprint arXiv:2305.13677}{}{}{}.
\PrintBackRefs{\CurrentBib}

\bibitem [\protect \citeauthoryear {%
MacAvaney%
\ \protect \BOthers {.}}{%
MacAvaney%
\ \protect \BOthers {.}}{%
{\protect \APACyear {2019}}%
}]{%
macavaney2019hate}
\APACinsertmetastar {%
macavaney2019hate}%
\begin{APACrefauthors}%
MacAvaney, S.%
\BCBT {}\ \BOthersPeriod {.}
\end{APACrefauthors}%
\unskip\
\newblock
\APACrefYearMonthDay{2019}{}{}.
\newblock
{\BBOQ}\APACrefatitle {Hate speech detection: Challenges and solutions} {Hate
  speech detection: Challenges and solutions}.{\BBCQ}
\newblock
\APACjournalVolNumPages{PloS one}{14}{8}{e0221152}.
\PrintBackRefs{\CurrentBib}

\bibitem [\protect \citeauthoryear {%
Meng%
\ \protect \BOthers {.}}{%
Meng%
\ \protect \BOthers {.}}{%
{\protect \APACyear {2020}}%
}]{%
meng2020text}
\APACinsertmetastar {%
meng2020text}%
\begin{APACrefauthors}%
Meng, Y.%
\BCBT {}\ \BOthersPeriod {.}
\end{APACrefauthors}%
\unskip\
\newblock
\APACrefYearMonthDay{2020}{}{}.
\newblock
{\BBOQ}\APACrefatitle {Text classification using label names only: A language
  model self-training approach} {Text classification using label names only: A
  language model self-training approach}.{\BBCQ}
\newblock
\APACjournalVolNumPages{arXiv preprint arXiv:2010.07245}{}{}{}.
\PrintBackRefs{\CurrentBib}

\bibitem [\protect \citeauthoryear {%
Min%
\ \protect \BOthers {.}}{%
Min%
\ \protect \BOthers {.}}{%
{\protect \APACyear {2023}}%
}]{%
min2023recent}
\APACinsertmetastar {%
min2023recent}%
\begin{APACrefauthors}%
Min, B.%
\BCBT {}\ \BOthersPeriod {.}
\end{APACrefauthors}%
\unskip\
\newblock
\APACrefYearMonthDay{2023}{}{}.
\newblock
{\BBOQ}\APACrefatitle {Recent advances in natural language processing via large
  pre-trained language models: A survey} {Recent advances in natural language
  processing via large pre-trained language models: A survey}.{\BBCQ}
\newblock
\APACjournalVolNumPages{ACM Computing Surveys}{56}{2}{1--40}.
\PrintBackRefs{\CurrentBib}

\bibitem [\protect \citeauthoryear {%
Muennighoff%
\ \protect \BOthers {.}}{%
Muennighoff%
\ \protect \BOthers {.}}{%
{\protect \APACyear {2022}}%
}]{%
muennighoff2022crosslingual}
\APACinsertmetastar {%
muennighoff2022crosslingual}%
\begin{APACrefauthors}%
Muennighoff, N.%
\BCBT {}\ \BOthersPeriod {.}
\end{APACrefauthors}%
\unskip\
\newblock
\APACrefYearMonthDay{2022}{}{}.
\newblock
{\BBOQ}\APACrefatitle {Crosslingual generalization through multitask
  finetuning} {Crosslingual generalization through multitask
  finetuning}.{\BBCQ}
\newblock
\APACjournalVolNumPages{arXiv preprint arXiv:2211.01786}{}{}{}.
\PrintBackRefs{\CurrentBib}

\bibitem [\protect \citeauthoryear {%
Oliveira%
\ \protect \BOthers {.}}{%
Oliveira%
\ \protect \BOthers {.}}{%
{\protect \APACyear {2023}}%
}]{%
oliveira2023good}
\APACinsertmetastar {%
oliveira2023good}%
\begin{APACrefauthors}%
Oliveira, A\BPBI S.%
\BCBT {}\ \BOthersPeriod {.}
\end{APACrefauthors}%
\unskip\
\newblock
\APACrefYearMonthDay{2023}{}{}.
\newblock
{\BBOQ}\APACrefatitle {How Good Is ChatGPT For Detecting Hate Speech In
  Portuguese?} {How good is chatgpt for detecting hate speech in
  portuguese?}{\BBCQ}
\newblock
\BIn{} \APACrefbtitle {Anais do XIV Simp{\'o}sio Brasileiro de Tecnologia da
  Informa{\c{c}}{\~a}o e da Linguagem Humana} {Anais do xiv simp{\'o}sio
  brasileiro de tecnologia da informa{\c{c}}{\~a}o e da linguagem humana}\
  (\BPGS\ 94--103).
\PrintBackRefs{\CurrentBib}

\bibitem [\protect \citeauthoryear {%
OpenAI%
}{%
OpenAI%
}{%
{\protect \APACyear {2023}}%
}]{%
openai2023gpt}
\APACinsertmetastar {%
openai2023gpt}%
\begin{APACrefauthors}%
OpenAI, R.%
\end{APACrefauthors}%
\unskip\
\newblock
\APACrefYearMonthDay{2023}{}{}.
\newblock
{\BBOQ}\APACrefatitle {GPT-4 technical report} {Gpt-4 technical report}.{\BBCQ}
\newblock
\APACjournalVolNumPages{arXiv}{}{}{2303--08774}.
\PrintBackRefs{\CurrentBib}

\bibitem [\protect \citeauthoryear {%
Peters%
\ \protect \BOthers {.}}{%
Peters%
\ \protect \BOthers {.}}{%
{\protect \APACyear {1802}}%
}]{%
peters1802deep}
\APACinsertmetastar {%
peters1802deep}%
\begin{APACrefauthors}%
Peters, M\BPBI E.%
\BCBT {}\ \BOthersPeriod {.}
\end{APACrefauthors}%
\unskip\
\newblock
\APACrefYearMonthDay{1802}{}{}.
\newblock
{\BBOQ}\APACrefatitle {Deep contextualized word representations. CoRR
  abs/1802.05365 (2018)} {Deep contextualized word representations. corr
  abs/1802.05365 (2018)}.{\BBCQ}
\newblock
\APACjournalVolNumPages{arXiv preprint arXiv:1802.05365}{}{}{}.
\PrintBackRefs{\CurrentBib}

\bibitem [\protect \citeauthoryear {%
Pham%
\ \protect \BOthers {.}}{%
Pham%
\ \protect \BOthers {.}}{%
{\protect \APACyear {2020}}%
}]{%
pham2020universal}
\APACinsertmetastar {%
pham2020universal}%
\begin{APACrefauthors}%
Pham, Q\BPBI H.%
\BCBT {}\ \BOthersPeriod {.}
\end{APACrefauthors}%
\unskip\
\newblock
\APACrefYearMonthDay{2020}{}{}.
\newblock
{\BBOQ}\APACrefatitle {From universal language model to downstream task:
  Improving RoBERTa-based Vietnamese hate speech detection} {From universal
  language model to downstream task: Improving roberta-based vietnamese hate
  speech detection}.{\BBCQ}
\newblock
\BIn{} \APACrefbtitle {2020 12th International Conference on Knowledge and
  Systems Engineering (KSE)} {2020 12th international conference on knowledge
  and systems engineering (kse)}\ (\BPGS\ 37--42).
\PrintBackRefs{\CurrentBib}

\bibitem [\protect \citeauthoryear {%
Plaza-del Arco%
\ \protect \BOthers {.}}{%
Plaza-del Arco%
\ \protect \BOthers {.}}{%
{\protect \APACyear {2021}}%
}]{%
plaza2021comparing}
\APACinsertmetastar {%
plaza2021comparing}%
\begin{APACrefauthors}%
Plaza-del Arco, F\BPBI M.%
\BCBT {}\ \BOthersPeriod {.}
\end{APACrefauthors}%
\unskip\
\newblock
\APACrefYearMonthDay{2021}{}{}.
\newblock
{\BBOQ}\APACrefatitle {Comparing pre-trained language models for Spanish hate
  speech detection} {Comparing pre-trained language models for spanish hate
  speech detection}.{\BBCQ}
\newblock
\APACjournalVolNumPages{Expert Systems with Applications}{166}{}{114120}.
\PrintBackRefs{\CurrentBib}

\bibitem [\protect \citeauthoryear {%
Poletto%
\ \protect \BOthers {.}}{%
Poletto%
\ \protect \BOthers {.}}{%
{\protect \APACyear {2021}}%
}]{%
poletto2021resources}
\APACinsertmetastar {%
poletto2021resources}%
\begin{APACrefauthors}%
Poletto, F.%
\BCBT {}\ \BOthersPeriod {.}
\end{APACrefauthors}%
\unskip\
\newblock
\APACrefYearMonthDay{2021}{}{}.
\newblock
{\BBOQ}\APACrefatitle {Resources and benchmark corpora for hate speech
  detection: a systematic review} {Resources and benchmark corpora for hate
  speech detection: a systematic review}.{\BBCQ}
\newblock
\APACjournalVolNumPages{Language Resources and Evaluation}{55}{}{477--523}.
\PrintBackRefs{\CurrentBib}

\bibitem [\protect \citeauthoryear {%
Rathnayake%
\ \protect \BOthers {.}}{%
Rathnayake%
\ \protect \BOthers {.}}{%
{\protect \APACyear {2022}}%
}]{%
rathnayake2022adapter}
\APACinsertmetastar {%
rathnayake2022adapter}%
\begin{APACrefauthors}%
Rathnayake, H.%
\BCBT {}\ \BOthersPeriod {.}
\end{APACrefauthors}%
\unskip\
\newblock
\APACrefYearMonthDay{2022}{}{}.
\newblock
{\BBOQ}\APACrefatitle {Adapter-based fine-tuning of pre-trained multilingual
  language models for code-mixed and code-switched text classification}
  {Adapter-based fine-tuning of pre-trained multilingual language models for
  code-mixed and code-switched text classification}.{\BBCQ}
\newblock
\APACjournalVolNumPages{Knowledge and Information Systems}{64}{7}{1937--1966}.
\PrintBackRefs{\CurrentBib}

\bibitem [\protect \citeauthoryear {%
Reiss%
}{%
Reiss%
}{%
{\protect \APACyear {2023}}%
}]{%
reiss2023testing}
\APACinsertmetastar {%
reiss2023testing}%
\begin{APACrefauthors}%
Reiss, M\BPBI V.%
\end{APACrefauthors}%
\unskip\
\newblock
\APACrefYearMonthDay{2023}{}{}.
\newblock
{\BBOQ}\APACrefatitle {Testing the reliability of chatgpt for text annotation
  and classification: A cautionary remark} {Testing the reliability of chatgpt
  for text annotation and classification: A cautionary remark}.{\BBCQ}
\newblock
\APACjournalVolNumPages{arXiv preprint arXiv:2304.11085}{}{}{}.
\PrintBackRefs{\CurrentBib}

\bibitem [\protect \citeauthoryear {%
R{\"o}ttger%
\ \protect \BOthers {.}}{%
R{\"o}ttger%
\ \protect \BOthers {.}}{%
{\protect \APACyear {2021}}%
}]{%
rottger-etal-2021-hatecheck}
\APACinsertmetastar {%
rottger-etal-2021-hatecheck}%
\begin{APACrefauthors}%
R{\"o}ttger, P.%
\BCBT {}\ \BOthersPeriod {.}
\end{APACrefauthors}%
\unskip\
\newblock
\APACrefYearMonthDay{2021}{{\APACmonth{08}}}{}.
\newblock
{\BBOQ}\APACrefatitle {{H}ate{C}heck: Functional Tests for Hate Speech
  Detection Models} {{H}ate{C}heck: Functional tests for hate speech detection
  models}.{\BBCQ}
\newblock
\BIn{} \APACrefbtitle {Proceedings of the 59th Annual Meeting of the
  Association for Computational Linguistics and the 11th International Joint
  Conference on Natural Language Processing (Volume 1: Long Papers)}
  {Proceedings of the 59th annual meeting of the association for computational
  linguistics and the 11th international joint conference on natural language
  processing (volume 1: Long papers)}\ (\BPGS\ 41--58).
\newblock
\APACaddressPublisher{Online}{Association for Computational Linguistics}.
\newblock
\begin{APACrefURL} \url{https://aclanthology.org/2021.acl-long.4}
  \end{APACrefURL}
\newblock
\begin{APACrefDOI} \doi{10.18653/v1/2021.acl-long.4} \end{APACrefDOI}
\PrintBackRefs{\CurrentBib}

\bibitem [\protect \citeauthoryear {%
Schmidt%
\ \BBA {} Wiegand%
}{%
Schmidt%
\ \BBA {} Wiegand%
}{%
{\protect \APACyear {2017}}%
}]{%
schmidt2017survey}
\APACinsertmetastar {%
schmidt2017survey}%
\begin{APACrefauthors}%
Schmidt, A.%
\BCBT {}\ \BBA {} Wiegand, M.%
\end{APACrefauthors}%
\unskip\
\newblock
\APACrefYearMonthDay{2017}{}{}.
\newblock
{\BBOQ}\APACrefatitle {A survey on hate speech detection using natural language
  processing} {A survey on hate speech detection using natural language
  processing}.{\BBCQ}
\newblock
\BIn{} \APACrefbtitle {Proceedings of the fifth international workshop on
  natural language processing for social media} {Proceedings of the fifth
  international workshop on natural language processing for social media}\
  (\BPGS\ 1--10).
\PrintBackRefs{\CurrentBib}

\bibitem [\protect \citeauthoryear {%
Sheth%
\ \protect \BOthers {.}}{%
Sheth%
\ \protect \BOthers {.}}{%
{\protect \APACyear {2023}}%
}]{%
sheth2023peace}
\APACinsertmetastar {%
sheth2023peace}%
\begin{APACrefauthors}%
Sheth, P.%
\BCBT {}\ \BOthersPeriod {.}
\end{APACrefauthors}%
\unskip\
\newblock
\APACrefYearMonthDay{2023}{}{}.
\newblock
{\BBOQ}\APACrefatitle {Peace: Cross-platform hate speech detection-a
  causality-guided framework} {Peace: Cross-platform hate speech detection-a
  causality-guided framework}.{\BBCQ}
\newblock
\BIn{} \APACrefbtitle {Joint European Conference on Machine Learning and
  Knowledge Discovery in Databases} {Joint european conference on machine
  learning and knowledge discovery in databases}\ (\BPGS\ 559--575).
\PrintBackRefs{\CurrentBib}

\bibitem [\protect \citeauthoryear {%
Sheth%
\ \protect \BOthers {.}}{%
Sheth%
\ \protect \BOthers {.}}{%
{\protect \APACyear {2024}}%
}]{%
sheth2024causality}
\APACinsertmetastar {%
sheth2024causality}%
\begin{APACrefauthors}%
Sheth, P.%
\BCBT {}\ \BOthersPeriod {.}
\end{APACrefauthors}%
\unskip\
\newblock
\APACrefYearMonthDay{2024}{}{}.
\newblock
{\BBOQ}\APACrefatitle {Causality Guided Disentanglement for Cross-Platform Hate
  Speech Detection} {Causality guided disentanglement for cross-platform hate
  speech detection}.{\BBCQ}
\newblock
\BIn{} \APACrefbtitle {Proceedings of the 17th ACM International Conference on
  Web Search and Data Mining} {Proceedings of the 17th acm international
  conference on web search and data mining}\ (\BPGS\ 626--635).
\PrintBackRefs{\CurrentBib}

\bibitem [\protect \citeauthoryear {%
Sohn%
\ \BBA {} Lee%
}{%
Sohn%
\ \BBA {} Lee%
}{%
{\protect \APACyear {2019}}%
}]{%
sohn2019mc}
\APACinsertmetastar {%
sohn2019mc}%
\begin{APACrefauthors}%
Sohn, H.%
\BCBT {}\ \BBA {} Lee, H.%
\end{APACrefauthors}%
\unskip\
\newblock
\APACrefYearMonthDay{2019}{}{}.
\newblock
{\BBOQ}\APACrefatitle {Mc-bert4hate: Hate speech detection using multi-channel
  bert for different languages and translations} {Mc-bert4hate: Hate speech
  detection using multi-channel bert for different languages and
  translations}.{\BBCQ}
\newblock
\BIn{} \APACrefbtitle {2019 International Conference on Data Mining Workshops
  (ICDMW)} {2019 international conference on data mining workshops (icdmw)}\
  (\BPGS\ 551--559).
\PrintBackRefs{\CurrentBib}

\bibitem [\protect \citeauthoryear {%
Stappen%
, Brunn%
\BCBL {}\ \BBA {} Schuller%
}{%
Stappen%
\ \protect \BOthers {.}}{%
{\protect \APACyear {2020}}%
}]{%
stappen2020cross}
\APACinsertmetastar {%
stappen2020cross}%
\begin{APACrefauthors}%
Stappen, L.%
, Brunn, F.%
\BCBL {}\ \BBA {} Schuller, B.%
\end{APACrefauthors}%
\unskip\
\newblock
\APACrefYearMonthDay{2020}{}{}.
\newblock
{\BBOQ}\APACrefatitle {Cross-lingual zero-and few-shot hate speech detection
  utilising frozen transformer language models and AXEL} {Cross-lingual
  zero-and few-shot hate speech detection utilising frozen transformer language
  models and axel}.{\BBCQ}
\newblock
\APACjournalVolNumPages{arXiv preprint arXiv:2004.13850}{}{}{}.
\PrintBackRefs{\CurrentBib}

\bibitem [\protect \citeauthoryear {%
Sun%
\ \protect \BOthers {.}}{%
Sun%
\ \protect \BOthers {.}}{%
{\protect \APACyear {2019}}%
}]{%
sun2019fine}
\APACinsertmetastar {%
sun2019fine}%
\begin{APACrefauthors}%
Sun, C.%
\BCBT {}\ \BOthersPeriod {.}
\end{APACrefauthors}%
\unskip\
\newblock
\APACrefYearMonthDay{2019}{}{}.
\newblock
{\BBOQ}\APACrefatitle {How to fine-tune bert for text classification?} {How to
  fine-tune bert for text classification?}{\BBCQ}
\newblock
\BIn{} \APACrefbtitle {Chinese Computational Linguistics: 18th China National
  Conference, CCL 2019, Kunming, China, October 18--20, 2019, Proceedings 18}
  {Chinese computational linguistics: 18th china national conference, ccl 2019,
  kunming, china, october 18--20, 2019, proceedings 18}\ (\BPGS\ 194--206).
\PrintBackRefs{\CurrentBib}

\bibitem [\protect \citeauthoryear {%
Touvron%
\ \protect \BOthers {.}}{%
Touvron%
\ \protect \BOthers {.}}{%
{\protect \APACyear {2023}}%
}]{%
touvron2023llama}
\APACinsertmetastar {%
touvron2023llama}%
\begin{APACrefauthors}%
Touvron, H.%
\BCBT {}\ \BOthersPeriod {.}
\end{APACrefauthors}%
\unskip\
\newblock
\APACrefYearMonthDay{2023}{}{}.
\newblock
{\BBOQ}\APACrefatitle {Llama 2: Open foundation and fine-tuned chat models}
  {Llama 2: Open foundation and fine-tuned chat models}.{\BBCQ}
\newblock
\APACjournalVolNumPages{arXiv preprint arXiv:2307.09288}{}{}{}.
\PrintBackRefs{\CurrentBib}

\bibitem [\protect \citeauthoryear {%
Velankar%
, Patil%
\BCBL {}\ \BBA {} Joshi%
}{%
Velankar%
\ \protect \BOthers {.}}{%
{\protect \APACyear {2022}}%
}]{%
velankar2022mono}
\APACinsertmetastar {%
velankar2022mono}%
\begin{APACrefauthors}%
Velankar, A.%
, Patil, H.%
\BCBL {}\ \BBA {} Joshi, R.%
\end{APACrefauthors}%
\unskip\
\newblock
\APACrefYearMonthDay{2022}{}{}.
\newblock
{\BBOQ}\APACrefatitle {Mono vs multilingual bert for hate speech detection and
  text classification: A case study in marathi} {Mono vs multilingual bert for
  hate speech detection and text classification: A case study in
  marathi}.{\BBCQ}
\newblock
\BIn{} \APACrefbtitle {IAPR Workshop on Artificial Neural Networks in Pattern
  Recognition} {Iapr workshop on artificial neural networks in pattern
  recognition}\ (\BPGS\ 121--128).
\PrintBackRefs{\CurrentBib}

\bibitem [\protect \citeauthoryear {%
Zhang%
\ \protect \BOthers {.}}{%
Zhang%
\ \protect \BOthers {.}}{%
{\protect \APACyear {2021}}%
}]{%
zhang2021smedbert}
\APACinsertmetastar {%
zhang2021smedbert}%
\begin{APACrefauthors}%
Zhang, T.%
\BCBT {}\ \BOthersPeriod {.}
\end{APACrefauthors}%
\unskip\
\newblock
\APACrefYearMonthDay{2021}{}{}.
\newblock
{\BBOQ}\APACrefatitle {SMedBERT: A knowledge-enhanced pre-trained language
  model with structured semantics for medical text mining} {Smedbert: A
  knowledge-enhanced pre-trained language model with structured semantics for
  medical text mining}.{\BBCQ}
\newblock
\APACjournalVolNumPages{arXiv preprint arXiv:2108.08983}{}{}{}.
\PrintBackRefs{\CurrentBib}

\bibitem [\protect \citeauthoryear {%
Zhao%
, Zhang%
\BCBL {}\ \BBA {} Hopfgartner%
}{%
Zhao%
\ \protect \BOthers {.}}{%
{\protect \APACyear {2021}}%
}]{%
zhao2021comparative}
\APACinsertmetastar {%
zhao2021comparative}%
\begin{APACrefauthors}%
Zhao, Z.%
, Zhang, Z.%
\BCBL {}\ \BBA {} Hopfgartner, F.%
\end{APACrefauthors}%
\unskip\
\newblock
\APACrefYearMonthDay{2021}{}{}.
\newblock
{\BBOQ}\APACrefatitle {A comparative study of using pre-trained language models
  for toxic comment classification} {A comparative study of using pre-trained
  language models for toxic comment classification}.{\BBCQ}
\newblock
\BIn{} \APACrefbtitle {Companion Proceedings of the Web Conference 2021}
  {Companion proceedings of the web conference 2021}\ (\BPGS\ 500--507).
\PrintBackRefs{\CurrentBib}

\bibitem [\protect \citeauthoryear {%
Zhu%
\ \protect \BOthers {.}}{%
Zhu%
\ \protect \BOthers {.}}{%
{\protect \APACyear {2023}}%
}]{%
zhu2023can}
\APACinsertmetastar {%
zhu2023can}%
\begin{APACrefauthors}%
Zhu, Y.%
\BCBT {}\ \BOthersPeriod {.}
\end{APACrefauthors}%
\unskip\
\newblock
\APACrefYearMonthDay{2023}{}{}.
\newblock
{\BBOQ}\APACrefatitle {Can chatgpt reproduce human-generated labels? a study of
  social computing tasks} {Can chatgpt reproduce human-generated labels? a
  study of social computing tasks}.{\BBCQ}
\newblock
\APACjournalVolNumPages{arXiv preprint arXiv:2304.10145}{}{}{}.
\PrintBackRefs{\CurrentBib}

\bibitem [\protect \citeauthoryear {%
Zia%
\ \protect \BOthers {.}}{%
Zia%
\ \protect \BOthers {.}}{%
{\protect \APACyear {2022}}%
}]{%
zia2022improving}
\APACinsertmetastar {%
zia2022improving}%
\begin{APACrefauthors}%
Zia, H\BPBI B.%
\BCBT {}\ \BOthersPeriod {.}
\end{APACrefauthors}%
\unskip\
\newblock
\APACrefYearMonthDay{2022}{}{}.
\newblock
{\BBOQ}\APACrefatitle {Improving zero-shot cross-lingual hate speech detection
  with pseudo-label fine-tuning of transformer language models} {Improving
  zero-shot cross-lingual hate speech detection with pseudo-label fine-tuning
  of transformer language models}.{\BBCQ}
\newblock
\BIn{} \APACrefbtitle {Proceedings of the International AAAI conference on web
  and social media} {Proceedings of the international aaai conference on web
  and social media}\ (\BVOL~16, \BPGS\ 1435--1439).
\PrintBackRefs{\CurrentBib}

\end{thebibliography}
\end{document}